\documentclass{article}

 \usepackage[preprint]{neurips_2026}

\usepackage[utf8]{inputenc} 
\usepackage[T1]{fontenc}    
\usepackage{hyperref}       
\usepackage{url}            
\usepackage{booktabs}       
\usepackage{amsfonts}       
\usepackage{nicefrac}       
\usepackage{microtype}      
\usepackage{adjustbox}
\usepackage{subcaption}
\usepackage{custom_commands}

\title{ChronoEarth-492K: A Large Scale and Long Horizon Spatiotemporal Hyperspectral Earth Observation Dataset and Benchmark}

\author{%
  Haozhe Si\textsuperscript{1}, 
  Yuxuan Wan\textsuperscript{2}, 
  Yuqing Wang\textsuperscript{2}, 
  Minh Do\textsuperscript{1},
  Han Zhao\textsuperscript{2,}\thanks{All correspondence should be addressed to Han Zhao: \texttt{hanzhao@illinois.edu}}\\
  \textsuperscript{1}Department of Electrical and Computer Engineering\\
  \textsuperscript{2}Siebel School of Computing and Data Science\\
  University of Illinois Urbana-Champaign, IL, USA \\
}

\begin{document}

\maketitle
\begin{abstract}
    Hyperspectral imaging (HSI) provides dense spectral information for the Earth’s surface, enabling material-level understanding of land cover and ecosystem dynamics. Despite recent progress in hyperspectral self-supervised learning (SSL), existing datasets remain temporally shallow, limiting the development of long-horizon spatiotemporal modeling. To address this gap, we introduce ChronoEarth-492K, the first large-scale, temporally calibrated hyperspectral SSL dataset built upon NASA’s EO-1 Hyperion mission, the world’s longest continuous hyperspectral archive up to date (2001–2017). ChronoEarth-492K comprises 492,354 radiometrically harmonized patches across 185,398 global locations over 17 years, with 28,786 sites containing multi-temporal sequences ($\geq$ 3 observations) that enable both short- and long-horizon temporal analysis. Building on this foundation, we establish the ChronoEarth-Benchmark, a unified evaluation suite spanning static, short-horizon, and long-horizon temporal tasks, constructed from six open-source geospatial products covering land cover, crop type, forest dynamics, and soil properties. We further introduce a standardized evaluation protocol and report extensive baseline results across state-of-the-art hyperspectral foundation models. Together, ChronoEarth and benchmark provide the first large-scale, temporally grounded platform for systematic spatiotemporal hyperspectral representation learning. The code and latest version of the dataset are available through the project page: \texttt{https://uiuctml.github.io/ChronoEarth492K/}.
\end{abstract}

\section{Introduction}
\label{sec:intro}
Earth observation (EO) is essential for understanding global environmental dynamics, monitoring human activity impacts, and guiding sustainable development. Among various remote sensing modalities, hyperspectral imaging (HSI) acquires hundreds of contiguous spectral bands for each pixel, providing rich spectral–spatial information that enables material-level characterization of the Earth’s surface~\citep{hsi}. This fine spectral granularity enables precise discrimination of vegetation species, soil composition, and agricultural conditions, exceeding the descriptive capacity of multispectral sensors that typically contain fewer than twenty bands. Over the past two decades, spaceborne HSI missions~\citep{EO1H,PRISMA, EnMAP, gaofen5} have established orbital hyperspectral observation at tens-of-meters resolution with global coverage. As these archives expand in volume and temporal continuity, they increasingly support analysis beyond spatial–spectral representation toward temporal modeling, enabling the study of deforestation, land-use transitions, and long-term environmental change. 

Despite the rich spectral capacity of HSI, existing large-scale datasets are predominantly designed for static scene understanding rather than spatiotemporal modeling. Recent works~\citep{hyspecnet, msst, hsihybrid, spectralearth} have leveraged unlabeled HSI, primarily from EnMAP~\citep{EnMAP}, to develop self-supervised geospatial foundation models. While these efforts successfully capture global spatial variability, the limited temporal span of current HSI corpora restricts their ability to model long-term surface dynamics and, more importantly, prevents systematic evaluation of temporal generalization and forecasting capabilities. Consequently, spatiotemporal hyperspectral foundation modeling remains underexplored.

\begin{figure}[t]
    \centering
    \includegraphics[width=0.6\linewidth]{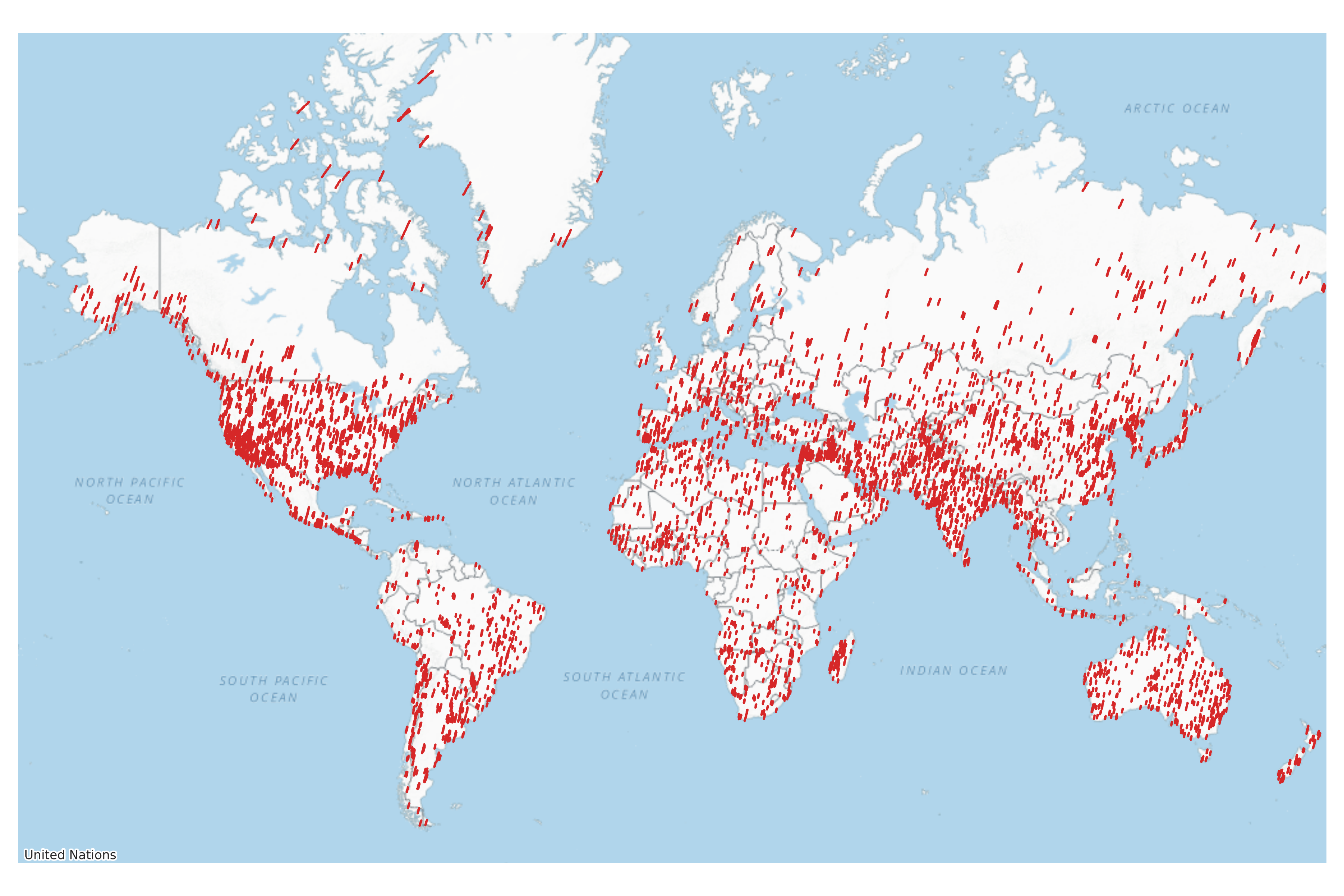}
    \caption{Global Distribution of the ChronoEarth dataset. The map illustrates the world-wide coverage of hyperspectral observations included in our dataset, highlighting its extensive spatial diversity across major continental regions.}
    \label{fig:global_dist}
\end{figure}

To address this limitation, we introduce ChronoEarth-492K, a large-scale, temporally calibrated hyperspectral dataset built upon NASA’s EO-1 Hyperion mission (2001–2017)~\citep{EO1H}, the longest continuous spaceborne hyperspectral archive to date. ChronoEarth-492K comprises 492,354 radiometrically harmonized patches across 185,398 global locations (\Cref{fig:global_dist}), with 28,786 locations containing multi-temporal sequences ($\geq$ 3 frames). This temporal depth enables both short-horizon aggregation and long-horizon prediction, providing a foundation for systematic spatiotemporal representation learning.

Building on this dataset, we introduce ChronoEarth-Benchmark, a unified evaluation suite for systematic assessment of spatial representation learning and temporal modeling. The benchmark integrates six open-source geospatial products, including land-cover~\citep{nlcd, clcd}, crop~\citep{corine, cdl}, forest~\citep{gfc}, and soil datasets~\citep{isda}. These datasets are carefully aligned with ChronoEarth observations. We define static, short-horizon, and long-horizon temporal tasks, enabling systematic evaluation of both spatial representation learning and temporal modeling. In addition, we design spatial-temporal and continental generalization settings, assessing the model robustness under domain shifts. Together, these components establish the first standardized platform for developing and evaluating temporally aware hyperspectral foundation models. This design enables controlled analysis of temporal aggregation, forecasting, and generalization under realistic distribution shifts.

We evaluate a range of state-of-the-art hyperspectral foundation models~\citep{spectralearth,dofa,satmae,hypersigma,less} on both static and temporal tasks. For temporal evaluation, we extend static architectures with multiple temporal adaptation strategies, including learnable temporal modeling modules and non-parametric aggregation methods (e.g., max pooling), enabling consistent comparison across short-horizon and long-horizon settings. Extensive experiments demonstrate that ChronoEarth-Benchmark effectively differentiates model capabilities under varying temporal configurations and distribution shifts, demonstrating that the benchmark differentiates model capabilities under varying temporal configurations and distribution shifts. In summary, our contributions are threefold:
\begin{itemize}
    \item We present ChronoEarth-492K, the first large-scale hyperspectral SSL dataset offering 17 years of temporally calibrated global observations;

    \item We introduce ChronoEarth-Benchmark, a unified evaluation suite comprising static, short-horizon, and long-horizon temporal tasks, enabling systematic assessment of spatiotemporal hyperspectral representation learning;

    \item We provide extensive empirical evaluations of state-of-the-art hyperspectral foundation models, establishing strong reference baselines and demonstrating the effectiveness of temporal pretraining and modeling under diverse settings.
\end{itemize}

\section{Related Works}
\textbf{Geospatial Datasets.}~
Large-scale geospatial representation learning has been primarily driven by multispectral imagery (MSI), which captures reflected radiance in a limited set of discrete spectral bands (typically 10–13) from platforms such as Sentinel-2 and Landsat. A number of self-supervised learning (SSL) datasets and benchmarks~\citep{geobench,less} have been developed for MSI, covering both pretraining corpora~\citep{ssl4eo,seco,ben} and standardized downstream tasks across classification~\citep{ben, eurosat, so2sat} and segmentation~\citep{nlcd,cdl,spectralgpt,dfc}. Despite their success, MSI datasets typically provide only 10–13 spectral bands, limiting spectral granularity and material-level characterization.

Hyperspectral datasets provide substantially denser spectral sampling ($>$100 contiguous bands) and have recently been curated to support self-supervised representation learning~\citep{hyspecnet,msst,hsihybrid,hypersigma,toulouse,greenhyperspectra}. These datasets are primarily designed for static spatial representation learning, without sustained temporal coverage. SpectralEarth~\citep{spectralearth} is the only recent effort incorporating temporal observations, providing approximately two years of EnMAP~\citep{EnMAP} data and leveraging temporal views as augmentation for contrastive pretraining. However, its design does not explicitly support long-horizon temporal modeling or systematic spatiotemporal evaluation.

\begin{figure*}[t]
    \centering
    \includegraphics[width=0.8\linewidth]{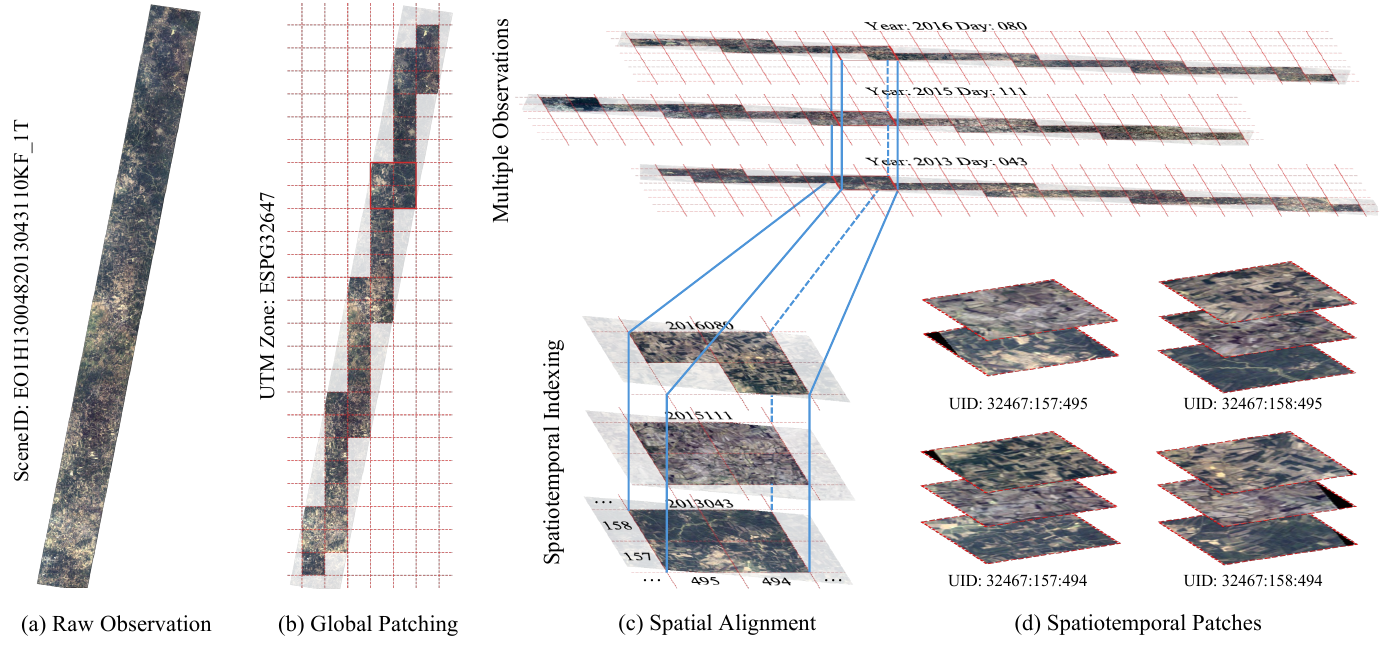}
    \caption{\textbf{Global Patching and Spatial Alignment for ChronoEarth.} \textbf{(a)} Raw EO-1 Hyperion observation.
    \textbf{(b)} Global patching using a UTM-zone–specific gridding system, where each patch is assigned a unique spatial identifier based on its zone and grid coordinates.
    \textbf{(c)} Spatial alignment of patches sharing the same identifier across multiple observations.
    \textbf{(d)} Formation of spatiotemporal patch sequences by aggregating aligned samples from different timestamps.}
    \label{fig:globalpatching}
\end{figure*}

\section{ChronoEarth-492K}
\textbf{Data Source.}~
ChronoEarth-492K is constructed from NASA’s EO-1 Hyperion mission. From all Level-1T terrain-corrected scenes available within this temporal span, we curate a globally distributed hyperspectral corpus designed specifically for large-scale spatiotemporal and hyperspectral self-supervised learning. To ensure spatial and radiometric consistency, we retained only acquisitions with reported cloud coverage below 10\%. To mitigate excessive oceanic coverage and balance geographic diversity, we manually defined nine continental-scale regions (Africa, Arctic, Europe, East Asia, Latin America, Oceania, North America, Southeast Asia, Southwest Asia), spanning the major inhabited landmasses, from which the final global collection was assembled.

\textbf{Spectral Harmonization.}~
Each Level-1T scene contains 242 spectral bands. Following previous works~\citep{process_eo1}, we remove unstable and low-signal bands (e.g., water absorption regions), retaining 155 spectrally consistent bands for all samples. All retained bands are aligned to a unified wavelength grid to ensure inter-scene compatibility across time and geography.

\textbf{Global Patching and Spatiotemporal Indexing.}~
To ensure deterministic spatial alignment across timestamps, each scene is resampled to a uniform 30m pixel size and divided into non-overlapping $128 \times 128$ pixel patches, aligned to the fixed global grid of its corresponding UTM zone. Each patch is assigned a unique location identifier (UID), which encodes its absolute grid location. Patches sharing the same UID but acquired at different timestamps form temporal sequences for the same geographic location. The global patching and alignment process is illustrated in Figure~\ref{fig:globalpatching}. Additional details provided in \Cref{appendix:chronoearth}.

\begin{figure*}[t]
    \centering
    \includegraphics[width=\linewidth]{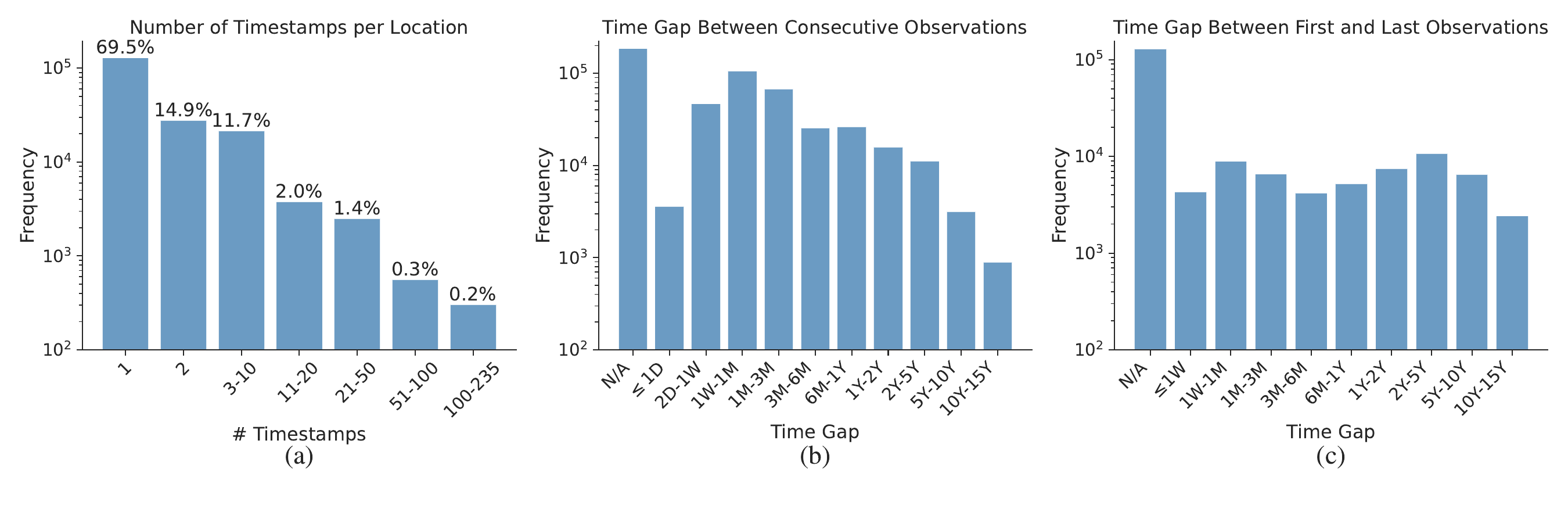}
    \vspace{-2em}
    \caption{Temporal distribution of ChronoEarth: (a) number of timestamps per location, (b) time gaps between consecutive acquisitions, and (c) temporal coverage per location. \textbf{N/A} refers to locations with single frame.}
    \label{fig:temporal_dist}
\end{figure*}

\textbf{Dataset Statistics}~
The ChronoEarth dataset comprises 492,354 hyperspectral patches of size $128 \times 128$, distributed across 185,398 unique spatial locations worldwide. Among these, 56,491 locations contain at least two observations, while 28,786 locations have three or more timestamps, forming multi-temporal sequences that enable temporal prediction beyond pairwise change detection. \Cref{fig:global_dist} shows the global spatial coverage, and \Cref{fig:temporal_dist} summarizes the temporal distribution. Most acquisition intervals are under one year, supporting short-term surface dynamics, while over one-third of multi-temporal locations span more than two years, enabling long-term land-cover analysis.

\begin{table}[t]
\centering
\caption{\textbf{Summary of Hyperspectral Remote Sensing Datasets for SSL.} ChronoEarth ranks second in spatial extent across existing hyperspectral remote-sensing datasets. Moreover, it spans the longest temporal range, providing the most extensive calibrated time-series coverage for SSL.}
\label{tab:ssl-compare}
\resizebox{\columnwidth}{!}{
\begin{tabular}{lcccccc}
\toprule
\textbf{Dataset} & \textbf{\# Img.}               & \textbf{Size}                      & \textbf{Sensor}    & \textbf{Bands} & \textbf{GSD} & \textbf{Temporal Coverage} \\ \midrule
HySpecNet-11k~\cite{hyspecnet}    & $\sim$11k                     & 128$\times$128                     & EnMAP              & 202            & 30m          & $\times$                   \\
MSST~\cite{msst}             & $\sim$20k                      & 64$\times$64                       & EnMAP              & 200            & 30m          & $\times$                   \\
HSIHybrid~\cite{hsihybrid}        & $\sim$4m & 9$\times$9                         & Multiple           & Variable       & --           & $\times$                   \\
HyperGlobal-450K~\cite{hypersigma} & $\sim$447k                    & 64$\times$64                       & EO-1 and Gaofen-5B & 242 and 330    & 30m          & $\times$                   \\
SpectralEarth~\cite{spectralearth}    & $\sim$538k                     & 128$\times$128                     & EnMAP              & 202            & 30m          & 2 Years                    \\ \midrule
ChronoEarth-492K       & $\sim$492k & 128$\times$128 & EO-1               & 155            & 30m          & 17 Years                   \\ \bottomrule
\end{tabular}
}
\end{table}

\textbf{Comparison with Existing Hyperspectral Datasets.}
Table~\ref{tab:ssl-compare} summarizes ChronoEarth in the context of prior hyperspectral SSL datasets. Compared to existing hyperspectral SSL datasets, ChronoEarth provides substantially longer temporal coverage (17 years), enabling systematic spatiotemporal representation learning.

\section{ChronoEarth-Benchmark}
To evaluate spatiotemporal hyperspectral representation learning at scale, we construct ChronoEarth-Benchmark by integrating seven publicly available geospatial products. These datasets span agriculture, land cover, soil properties, and forest change across multiple continents, enabling segmentation, multi-label classification, and change detection tasks under spatial distribution shifts. five of the six products provide multi-year annotations, allowing the ChronoEarth-Benchmark to emphasize temporal analysis. 

\subsection{Dataset Acquisition and Processing}
To fully leverage both the label layers and the hyperspectral observations, we first build a comprehensive dictionary of all spatial locations covered by the ChronoEarth dataset, indexed by UID and independent of acquisition time. Using this unified location reference, we recover the grid from the corresponding UTM zone for each label layer and match UIDs to extract all associated patches. As a result, a given label may correspond to multiple temporal observations in some years, or none in others, depending on data availability. Importantly, we retain all valid location–label pairs as potential supervision targets, enabling the construction of long-horizon temporal tasks.

\textbf{Label Processing and Filtering.}~
We do not rebalance class distributions, as the observed label frequencies reflect real-world geographic patterns. To reduce noise from extremely rare categories, labels with frequency below 1\% are grouped into a background class.

We further apply entropy-based filtering to remove low-diversity patches. Specifically, we compute the normalized Shannon entropy of each label patch and discard samples below a threshold ($\tau = 0.1$), retaining semantically informative regions for training. Additional details are provided in \Cref{appendix:filter}.

\textbf{Leak-Proof Split Generation.}~
To prevent spatial leakage, we generate train/val/test splits using a distance-aware grouping strategy based on EO-1 orbital swaths. Patches from the same observation are grouped into spatial units, and overlapping units are merged into connected components. Splits are assigned at the component level, ensuring non-overlapping and spatially distinct partitions. This design induces a spatial out-of-distribution (OOD) setting, where test samples come from geographically distinct regions unseen during training, making the benchmark more challenging than random or patch-level partitioning. Examples of the resulting splits are provided in \Cref{appendix:downstream_dataset}.

\begin{table}[t]
\caption{\textbf{ChronoEarth-Benchmark datasets and task configurations.} 
For each dataset, we report the geographic coverage, number of semantic classes, and the number of samples under different task configurations. ``--'' indicates that the corresponding task variant is not defined for that dataset.}
\label{table:downstream}
\centering
\setlength{\tabcolsep}{6pt}
\renewcommand{\arraystretch}{1.15}
\resizebox{0.9\columnwidth}{!}{
\begin{tabular}{lcccccl}
\toprule
\textbf{Dataset}                              & \textbf{Region} & \textbf{Classes} & \textbf{Static} & \textbf{SH} & \textbf{LH} & \multicolumn{1}{c}{\textbf{Task Type}} \\ \midrule
GFC~\citep{gfc}                                           & Global          & 1                & 5,509          &   --                    &  --                     & Forest Change Detection                \\
ISDASoil~\citep{isda}    & Africa          & 8                & 17,808          & 1,549                  & --                     & Multi-class Classification             \\
CDL~\citep{cdl}          & USA             & 12               & 10,162          & 1,762                  & 538                   & Crop Type Segmentation                \\
CORINE~\citep{corine}                                        & EU              & 19               & 19,774          & --                 & --                     & Multi-class Classification             \\
NLCD-S~\citep{nlcd} & USA             & 16               & 55,476          & 4,077                  & 1,812                 & Land Cover Segmentation                \\
CLCD~\citep{clcd}        & China           & 7                & 17,614          & 1,069                  & 460                   & Land Cover Segmentation                \\
 \bottomrule
\end{tabular}
}
\end{table}

\subsection{Benchmark Tasks} \label{sec:tasks}
We briefly summarize the datasets used to construct ChronoEarth-Benchmark in \Cref{table:downstream}. Detailed statistics and dataset-specific processing procedures are provided in \Cref{appendix:downstream_dataset}. ChronoEarth-Benchmark defines a unified task taxonomy spanning static prediction, short-horizon temporal aggregation, long-horizon forecasting, and structured generalization under spatial and temporal shifts.

\textbf{Static Tasks.}~
Each sample pairs a label map with a spatially aligned hyperspectral observation from the same year. The objective is to predict labels from a single input. We include segmentation (CDL, CLCD, NLCD-S) and multi-label classification (ISDASoil) tasks.

\textbf{Short-Horizon Temporal Tasks (SH).}
SH tasks extend the static setting by providing multiple temporally adjacent observations corresponding to the same label. Models aggregate short-term temporal context under fully observed supervision. We construct SH variants for CDL, CLCD, NLCD-S, and ISDASoil.

\textbf{Long-Horizon Temporal Tasks (LH).}
LH tasks formulate future prediction problems: observations from 2001 to year $n$ are used to predict labels at year $n+1$. This setting requires modeling temporal dynamics under partially observed supervision. We construct LH tasks for CDL, CLCD, and NLCD-S.

\emph{Temporal Protocol.}~Sequences are filtered with a minimum observation threshold of 2 to exclude effectively static samples. All retained data are than normalized to a fixed length $T$ via sub-sampling or zero-padding, ensuring that ablation studies on temporal context are conducted over an identical set of locations. During evaluation, temporal sequences are deterministically fixed to guarantee fair comparison across methods.

\textbf{Spatial-Temporal Generalization.}~ Using CORINE, we construct controlled distribution shifts by splitting data into temporal ID (2001–2012) and temporal OOD (2013–2017), followed by distance-aware split generation algorithm applied to each temporal group. This yields ID, spatial OOD, temporal OOD, and joint spatial–temporal OOD settings.

\textbf{Continental Generalization.}
Using GFC, we evaluate cross-continental transfer by training on data-rich regions (Europe, North America, East Asia) and evaluating on underrepresented regions (Africa, Latin America, Oceania, Southwest Asia), simulating the annotation imbalance scenario.


\section{Experiments}
In this section, we evaluate ChronoEarth-Benchmark with the goal of assessing whether large-scale spatiotemporal hyperspectral pretraining provides meaningful benefits for downstream learning. Specifically, we study (1) whether pretraining on ChronoEarth improves model initialization compared to supervised training, (2) whether learned representations generalize across sensors and geographic distributions, (3) whether temporal modeling provides additional gains beyond static representations, and (4) whether the proposed temporal benchmarks can effectively differentiate model capabilities.
\subsection{Baseline Architectures}
\textbf{Static Baselines.}
We consider representative hyperspectral foundation models that natively support hyperspectral inputs or can be adapted with minimal modification. Specifically, we include SpectralViT~\citep{spectralearth}, HyperSigma~\citep{hypersigma}, DOFA~\citep{dofa}, and LESSViT~\citep{less}. SatMAE~\citep{satmae} is adapted by extending its channel-grouping module, and DINOv3~\citep{dinov3} is included as a general-purpose ViT baseline with its patch embedding layer modified to accept 155 input channels. All models are instantiated at the ViT-Base scale for a fair comparison.

\textbf{Temporal Baselines.}
While existing hyperspectral foundation models are primarily designed for static inputs, our benchmark includes temporal settings that require aggregating multiple observations. To enable a simple and consistent temporal baseline, we adopt a two-stage design. In the first stage, an HSI backbone is pretrained using standard self-supervised learning. In the second stage, the pretrained encoder is frozen and a lightweight temporal module is attached to operate on token-level features. This design allows the model to handle sparse and non-uniform sampling without requiring fixed temporal intervals, while isolating temporal modeling from spatial representation learning.

\subsection{Pretrain Setup}
\textbf{Static Pretraining.}~We pretrain SpectralViT, and LESSViT on the ChronoEarth-492K dataset under the static configuration using all 492,354 samples. SpectralViT is trained in a self-supervised manner following the Masked Autoencoder (MAE) paradigm~\citep{mae}, with 75\% of spatial patches masked during pretraining. For LESSViT, we train it with their proposed Hyper-MAE with a 75\% spectral mask. All pretrained models are trained for 200 epochs using the AdamW optimizer~\citep{adamw}, with a learning rate of $1\times10^{-4}$, weight decay of $5\times10^{-2}$, cosine learning rate annealing, and a 10-epoch warm-up.

\textbf{Stage-Two Temporal Pretraining.}
We perform a second-stage self-supervised training on multi-temporal sequences to enable temporal modeling. This stage uses only locations with at least three timestamps (28,786 locations), enabling next-frame prediction. Given a sequence $\{x_1, \dots, x_T\}$, the model predicts $x_T$ from $\{x_1, \dots, x_{T-1}\}$ with causal masks. The predicted features are passed through a reconstruction decoder to recover the target in pixel space, optimized using an $\ell_2$ reconstruction loss. During this stage, the pretrained encoder is frozen, and only the temporal module and decoder are trained. We construct this stage using SpectralViT as the backbone and train for 50 epochs, with all other configurations identical to the static pretraining. Additional details is provided in \Cref{appendix:temporal}.

\subsection{Evaluation Setup}

\textbf{Task Heads.}
For downstream evaluation, we attach task-specific heads to each backbone. We use a linear classification head for multi-label classification and a UPerNet~\citep{upernet} decoder for semantic segmentation. These heads are applied consistently across all methods.

\textbf{Supervised Baseline.}
To evaluate the benefit of self-supervised pretraining, we train SpectralViT and LESSViT backbones with task-specific heads from random initialization on the static downstream tasks. This provides a controlled reference for measuring improvements from pretraining.

\textbf{External Baselines.}
For DOFA, HyperSigma, SatMAE, and DINOv3, we directly adopt their publicly released pretrained backbones and evaluate them under the same downstream protocols.

\textbf{Temporal Adaptation.}
Our goal is to provide a controlled evaluation of temporal information under a unified framework. Rather than introducing new temporal architectures, we adopt lightweight modules to isolate the effects of temporal aggregation and pretraining, enabling the benchmark to differentiate modeling strategies without confounding complexity. To extend static backbones to temporal inputs, we define three strategies:
\begin{enumerate}[label=(\arabic*)]
    \item \emph{Max Pooling:}~Each frame is independently encoded using a shared backbone, and frame-level features are aggregated by max pooling before being passed to the task head. This serves as a non-parametric baseline that tests whether temporal context alone provides gains without explicit temporal modeling.
    \item \emph{AttentionPool:}~We attach the lightweight temporal module on top of frame-level features and train it directly with downstream supervision. This setup evaluates whether learned temporal aggregation improves over fixed pooling under identical supervision.
    \item \emph{Temporal SSL:}~We initialize the temporal module using stage-two temporal pretraining and fine-tune the full pipeline on downstream tasks. This setting evaluates whether temporal self-supervised learning provides additional benefits beyond supervised temporal modeling.
\end{enumerate}
\emph{Remark.} For temporal adaptation beyond max pooling, we use SpectralViT as the backbone due to its simplicity and strong static performance, enabling a controlled study of temporal modeling. LESSViT incurs substantially higher computational cost when extended to temporal attention, and is therefore evaluated with max pooling only. Extending it to full temporal modeling is left for future work. HyperSigma is evaluated only on static tasks due to its non-standard architecture.

\textbf{Model Selection.}
For all experiments, we perform hyperparameter tuning and select the best checkpoint based on validation performance. Full training details are provided in Appendix~\ref{appendix:hparam}.

\subsection{Quantitative Results}
We present results on static tasks, generalization under distribution shifts, cross-satellite transfer, and temporal modeling in both short- and long-horizon settings.

\begin{table*}[t]
\centering
\caption{\textbf{Static and Generalization Results.}
(a) Performance of baseline methods on static downstream tasks.
(b) Performance on the CORINE dataset under in-distribution (ID), temporal (T-OOD), spatial (S-OOD), and joint spatial--temporal (ST-OOD) settings.
(c) Performance on the GFC dataset under ID and cross-continental OOD evaluation.
\emph{Sup.} denotes models trained from random initialization using supervised learning on downstream tasks without self-supervised pretraining.
Best results per column are highlighted in \textbf{bold}.}

\newcommand{\panelheight}{0.1\textheight}

\begin{subtable}[t]{0.33\textwidth}
\centering
\caption{\textbf{Static Task Results}}
\label{table:static}
\begin{adjustbox}{max height=\panelheight,max width=\linewidth,keepaspectratio,center}
\setlength{\tabcolsep}{6pt}
\renewcommand{\arraystretch}{1.15}
\begin{tabular}{c c c c c}
\toprule
\multirow{2}{*}{\textbf{Methods}} 
& \multicolumn{1}{c}{\textbf{CLCD}} 
& \multicolumn{1}{c}{\textbf{CDL}} 
& \multicolumn{1}{c}{\textbf{NLCD-S}} 
& \multicolumn{1}{c}{\textbf{ISDASoil}} \\

& mIoU$\uparrow$ & mIoU$\uparrow$  & mIoU$\uparrow$  & mAP$\uparrow$ \\
\midrule
DOFA        & 47.77 & 12.60  & 24.08 & 51.20   \\
HyperSigma   & 41.53 & 11.45 & 23.05 & 54.50   \\
DINOv3    & 47.10  & 11.65 & 24.46 & 50.90   \\
SatMAE      & 41.25 & 11.94 & 22.23 & 49.80 \\\midrule
\emph{Sup.} SpectralViT & 48.01 &	15.50	& 29.10 & \textbf{57.79}\\
SpectralViT & 53.29 & 20.87 & 30.80 & 57.70 \\\midrule
\emph{Sup.} LESSViT & 37.66	& 3.49 &	19.05 &	50.15\\
LESSViT        & \textbf{54.84} & \textbf{23.91} & \textbf{33.59} 
            & 56.02 \\

\bottomrule
\end{tabular}
\end{adjustbox}
\end{subtable}
\hfill
\begin{subtable}[t]{0.32\textwidth}
\centering
\caption{\textbf{Static CORINE Results}}
\label{table:corine}
\begin{adjustbox}{max height=\panelheight,max width=\linewidth,keepaspectratio,center}
\setlength{\tabcolsep}{6pt}
\renewcommand{\arraystretch}{1.15}
\begin{tabular}{ccccc}
\toprule
\multirow{2}{*}{\textbf{Methods}} & \multicolumn{4}{c}{\textbf{CORINE} (mAP$\uparrow$)} \\
 & \textbf{ID} & \textbf{T-OOD} & \textbf{S-OOD} & \textbf{ST-OOD} \\
\midrule
DOFA      & 70.08 & 62.77 & 57.85 & 51.80 \\
HyperSigma~   & 59.99 & 58.09 & 54.75 & 49.18 \\
DINOv3      & 58.93 & 54.88 & 50.80 & 46.00 \\
SatMAE     & 67.26 & 58.69 & 55.77 & 47.20 \\\midrule
\emph{Sup.} SpectralViT & 74.14	& 63.30	& 60.48	& 55.21\\
SpectralViT & 83.98 & \textbf{68.52} & \textbf{64.15} & 56.19 \\\midrule
\emph{Sup.} LESSViT  & 55.47 &	51.34 &	49.01	&47.20 \\
LESSViT        & \textbf{84.07} & 68.22 & 62.52 & \textbf{58.56} \\
\bottomrule
\end{tabular}
\end{adjustbox}
\end{subtable}
\hfill
\begin{subtable}[t]{0.32\textwidth}
\centering
\caption{\textbf{GFC Results}}
\label{table:gfc}
\begin{adjustbox}{max height=\panelheight,max width=\linewidth,keepaspectratio,center}
\setlength{\tabcolsep}{6pt}
\renewcommand{\arraystretch}{1.15}
\begin{tabular}{ccccc}
\toprule
\multirow{2}{*}{\textbf{Methods}} & \multicolumn{2}{c}{\textbf{ID}} & \multicolumn{2}{c}{\textbf{OOD}} \\
 & mIoU$\uparrow$ & F1$\uparrow$ & mIoU$\uparrow$ & F1$\uparrow$ \\
\midrule
DOFA& 16.99 & 29.05 & 12.77 & 22.65 \\
DINOv3      & 15.91 & 27.45 & 11.51 & 20.64 \\
SatMAE~      &  9.56 & 17.45 &  2.96 &  5.74 \\\midrule
\emph{Sup.} SpectralViT & 23.26 & 37.74 & 16.04 & 27.65 \\
SpectralViT & \textbf{29.90} & \textbf{46.04} & \textbf{19.38} & \textbf{32.47} \\\midrule
\emph{Sup.} LESSViT       & 0.00 & 0.00 & 0.00 & 0.00 \\
LESSViT       & 19.29 & 32.34 & 16.13 & 27.78 \\
\bottomrule
\end{tabular}
\end{adjustbox}
\end{subtable}

\end{table*}

\textbf{Static Tasks.}
Results on static tasks are shown in \Cref{table:static}. Self-supervised pretraining consistently improves performance over supervised counterparts across SpectralViT and LESSViT, demonstrating that ChronoEarth provides a strong initialization for downstream learning. The benchmark also differentiates model capacity: LESSViT achieves the best overall performance, indicating that architectures explicitly modeling spatial–spectral interactions can better exploit hyperspectral structure.

\textbf{Static Generalization Tasks.}~We evaluate model robustness under distribution shifts using CORINE (\Cref{table:corine}) and GFC (\Cref{table:gfc}). On CORINE, performance degrades progressively from ID to T-OOD, S-OOD, and ST-OOD, with spatial shift introducing the largest drop, indicating that geographic variability is the dominant factor limiting generalization. On GFC, all methods exhibit clear degradation from ID to OOD, reflecting the difficulty of cross-continental transfer under annotation imbalance and geographic shift. Interestingly, SpectralViT outperforms LESSViT in this setting. We attribute this to the sparsity of change signals in GFC, where effective learning relies on strong spatial aggregation rather than fine-grained spatial–spectral modeling. In contrast, architectures such as LESSViT, which explicitly model spatial–spectral interactions, require denser supervision to fully realize their advantages. This effect is further evidenced by the supervised LESSViT baseline, which fails to converge under sparse change supervision.

Overall, these generalization tasks demonstrate that ChronoEarth provides a principled and realistic evaluation framework, capturing both controlled distribution shifts and large-scale real-world transfer. This enables systematic analysis of model robustness across varying supervision regimes and geographic conditions, highlighting the challenges that arise in practical deployment.

\begin{table}[t]
\caption{\textbf{Cross-Satellite Generalization Results.}
Models pretrained on ChronoEarth (EO-1 Hyperion)  are fine-tuned and evaluated under the same protocol.
Results for \emph{SpectralEarth} are directly taken from \citet{spectralearth}, where models are both pretrained and evaluated on EnMAP data.
Best results per column are highlighted in \textbf{bold}.}

\centering
\setlength{\tabcolsep}{5pt}
\renewcommand{\arraystretch}{1.15}
\resizebox{0.6\linewidth}{!}{
\begin{tabular}{cccccc}
\toprule
\multirow{2}{*}{\textbf{Pretraining Dataset}} 
& \textbf{BDFORET} & \textbf{BNETD} & \textbf{EuroCrops} & \textbf{CORINE} & \textbf{CDL} \\
& mIoU$\uparrow$ & mIoU$\uparrow$ & mIoU$\uparrow$ & mAP$\uparrow$ & mIoU$\uparrow$ \\
\cmidrule{2-6}
SpectralEarth & 76.30 & \textbf{49.46} & 69.34 & 75.33 & \textbf{77.44} \\
\bluerow
ChronoEarth (\textbf{ours})   & \textbf{76.39} & 44.34 & \textbf{69.82} & \textbf{79.02} & 73.66 \\
\bottomrule
\end{tabular}
}
\label{tab:spectralearth}
\end{table}

\textbf{Cross-Satellite Generalization.}~\Cref{tab:spectralearth} evaluates transfer from ChronoEarth (EO-1 Hyperion) to downstream tasks constructed on EnMAP-based datasets~\citep{EnMAP, spectralearth}. Models pretrained on ChronoEarth achieve performance comparable to those trained directly on SpectralEarth, demonstrating effective transfer across sensors with different spectral characteristics. These results indicate that, despite originating from a decommissioned satellite, the long-term hyperspectral observations in ChronoEarth capture representations that generalize to newer missions.
\begin{table}[t]
\vspace{-1.5em}
\centering
\caption{\textbf{Short-Horizon Task Results.}
We compare single-frame input ($T=1$) and multi-frame input ($T \leq 4$) across short-horizon tasks.
Results evaluate how effectively models leverage additional temporal observations.
The best result for each metric is highlighted in \textbf{bold}.
\emph{Italic} values accompanied by $\downarrow$ indicate performance degradation when additional frames are introduced.
Unless otherwise specified, this formatting convention applies to all subsequent tables.}

\label{table:sh}
\setlength{\tabcolsep}{6pt}
\renewcommand{\arraystretch}{1.15}
\resizebox{0.75\columnwidth}{!}{
\begin{tabular}{c cc cc cc cc}
\toprule
\multirow{2}{*}{\textbf{Methods}}
& \multicolumn{2}{c}{\textbf{CLCD} (mIoU$\uparrow$)}
& \multicolumn{2}{c}{\textbf{CDL} (mIoU$\uparrow$)}
& \multicolumn{2}{c}{\textbf{NLCD-S} (mIoU$\uparrow$)}
& \multicolumn{2}{c}{\textbf{ISDASoil} (mAP$\uparrow$)}\\

& \textbf{T=1} & \textbf{T$\le$4}
& \textbf{T=1} & \textbf{T$\le$4}
& \textbf{T=1} & \textbf{T$\le$4}
& \textbf{T=1} & \textbf{T$\le$4} \\
\cmidrule(lr){2-3}\cmidrule(lr){4-5}\cmidrule(lr){6-7}\cmidrule(lr){8-9}

DOFA\footnotesize{\emph{max}}       
& 39.88 & \textit{39.76}$\downarrow$
& 12.68 & 14.88
& 14.96 & 18.25
& 52.80 & \textit{50.63}$\downarrow$ \\

DINOv3\footnotesize{\emph{max}}
& 40.39 & \textit{39.27}$\downarrow$
& 10.80 & 13.51
& 11.07 & 16.98
& 47.32 & 48.89 \\

SatMAE\footnotesize{\emph{max}}      
& 34.88 & 37.78
& 12.34 & 12.41
& 15.14 & 16.83
& 49.53 & 50.99 \\

LESSViT\footnotesize{\emph{max}} 
& \textbf{51.19} & \textbf{54.69}
& \textbf{24.80} & \textbf{26.15}
& \textbf{30.58} & \textbf{31.02}
& \textbf{57.33} & \textbf{58.04} 
\\\midrule

SpectralViT\footnotesize{\emph{max}}
& 43.37 & 44.55
& 16.47 & 20.07
& 19.90 & \textit{19.82}$\downarrow$
& 56.49 & 56.72 \\

SpectralViT\footnotesize{\emph{attention}}
& -- & 43.77
& -- & 18.97
& -- & 22.88
& -- & 48.98 \\

SpectralViT\footnotesize{\emph{temporal}}
& -- & 45.57
& -- & 22.64
& -- & 23.59
& -- & 54.14 \\

\bottomrule
\end{tabular}
}
\end{table}

\textbf{Short-Horizon Tasks.}
\Cref{table:sh} compares single-frame input ($T=1$) and multi-frame input ($T \leq 4$) across short-horizon datasets. Incorporating additional temporal observations generally improves performance, indicating that temporally adjacent frames provide complementary information beyond static features. LESSViT achieves the strongest overall performance under max pooling, consistent with its strong spatial–spectral representation capacity. This suggests that a sufficiently powerful static backbone can already benefit from temporal context through simple aggregation.

More importantly, comparing the three SpectralViT variants reveals consistent trends on segmentation tasks. The temporally pretrained model (\textit{temporal}) outperforms both the max-pooling (\textit{max}) and supervised attention-based (\textit{attention}) variants across most tasks. This demonstrates that (1) stage-two temporal self-supervised pretraining provides consistent improvements, and (2) learned temporal aggregation is more effective than naive pooling when controlling for the same backbone. These results indicate that the short-horizon benchmark can differentiate temporal modeling strategies under a controlled setting, where improvements arise from better utilization of additional temporal context rather than differences in spatial representation.

Overall, the short-horizon benchmark evaluates a model’s ability to leverage additional temporal context, where effective temporal modeling, particularly through self-supervised pretraining, provides consistent gains beyond strong static representations.

\textbf{Long-Horizon Tasks.}~
\Cref{table:LH} reports results under increasing temporal context ($T \leq 2$, $T \leq 4$, $T \leq 8$). Extending the temporal history generally improves performance on CLCD and NLCD-S, indicating that gradual land-cover transitions benefit from longer-term observations.

Comparing the SpectralViT variants further reveals that the temporally pretrained model (\emph{temporal}) consistently outperforms both the max-pooling (\emph{max}) and attention-based (\emph{attention}) variants. This confirms that temporal self-supervised pretraining remains effective in the long-horizon setting and that learned temporal modeling is necessary to capture extended temporal dependencies.

An exception is observed on CDL, where performance becomes less stable as the temporal horizon increases, with several methods showing degradation at $T \leq 8$. This suggests that for time-sensitive targets such as crop type, longer historical context does not necessarily provide additional useful signal for the aggregation strategies considered, and may introduce noise or outdated information.

In this setting, the long-horizon benchmark evaluates a model’s ability to predict future states from historical observations. Effective temporal modeling is therefore essential, particularly for tasks involving long-term dynamics.

\begin{table}[t]
\centering
\caption{\textbf{Long-Horizon Task Results.}
We evaluate future prediction under increasing temporal context ($T \leq 2$, $T \leq 4$, $T \leq 8$).
Results measure how effectively models utilize longer historical observations to predict future states.}
\setlength{\tabcolsep}{6pt}
\renewcommand{\arraystretch}{1.15}
\resizebox{0,75\columnwidth}{!}{
\begin{tabular}{c ccc ccc ccc}
\toprule
\multirow{2}{*}{\textbf{Methods}}
& \multicolumn{3}{c}{\textbf{CLCD} (mIoU $\uparrow$)}
& \multicolumn{3}{c}{\textbf{NLCD-S} (mIoU $\uparrow$)}
& \multicolumn{3}{c}{\textbf{CDL} (mIoU $\uparrow$)} \\

& \textbf{T$\le$2} & \textbf{T$\le$4} & \textbf{T$\le$8}
& \textbf{T$\le$2} & \textbf{T$\le$4} & \textbf{T$\le$8}
& \textbf{T$\le$2} & \textbf{T$\le$4} & \textbf{T$\le$8}\\
\cmidrule(lr){2-4}\cmidrule(lr){5-7}\cmidrule(lr){8-10}

DOFA\footnotesize{\emph{max}}   
&25.51 & 35.34 & 38.97
&18.76 & 19.51 & 22.64
& 6.02 & 10.27 & 10.44 \\

DINOv3\footnotesize{\emph{max}}   
& 20.26 & 33.72 & 37.04
& 18.14 & 20.34 & 21.83
& 5.25 & 9.35  & 11.22 \\

SatMAE\footnotesize{\emph{max}}      
& 20.00 & 31.25 & \textit{29.85}$\downarrow$
& 13.66 & 18.22 & 19.40
& 5.43 & 10.19 & 11.80 \\

LESSViT\footnotesize{\emph{max}}   
& 38.80 & 50.08 & 54.64
& 28.36 & 33.75 & 35.52
& 9.97 & \textbf{18.83} & \textit{\textbf{14.43}}$\downarrow$\\ \midrule

SpectralViT\footnotesize{\emph{max}}
& 35.88 & 40.70 & 49.25
& 25.67 & 28.44 & 30.11
& \textbf{11.95} & 13.60 & \textit{11.37}$\downarrow$\\

SpectralViT\footnotesize{\emph{attention}}
& 24.28 & 32.63 & 37.75
& 25.31 & 26.97 & 27.37
& 7.43 & \textit{7.12}$\downarrow$ & \textit{4.53}$\downarrow$ \\

SpectralViT\footnotesize{\emph{temporal}}
& \textbf{43.61} & \textbf{51.26} & \textbf{55.61}
& \textbf{37.34} & \textbf{39.69} & \textbf{40.76}
& 5.92 & 10.35 & \textit{9.55}$\downarrow$\\

\bottomrule
\end{tabular}
}
\label{table:LH}
\end{table}

\section{Conclusion}
We introduce ChronoEarth-492K, a large-scale hyperspectral dataset with long-term temporal coverage, together with a systematic pipeline for constructing spatiotemporal patch sequences from irregular satellite observations. Building on this dataset, we propose ChronoEarth-Benchmark, which defines static, short-horizon, and long-horizon tasks, along with controlled spatial and temporal generalization settings. Together, these contributions establish a unified framework for developing and evaluating spatiotemporal hyperspectral models. Our empirical results demonstrate that large-scale hyperspectral pretraining provides strong initialization for downstream tasks and that temporal modeling offers additional benefits, particularly under long-horizon prediction settings. 
Overall, the ChronoEarth-Benchmark provides a controlled and large-scale setting for systematically evaluating these effects across models, tasks, and distribution shifts.

A key limitation of the current work is that temporal modeling is implemented as an extension on top of pretrained spatial backbones, rather than being learned jointly in an end-to-end spatiotemporal framework. In addition, the temporal observations are sparse and irregular, which poses inherent challenges for learning consistent temporal representations.

These observations suggest a clear direction for future research: developing native spatiotemporal hyperspectral foundation models that integrate spatial, spectral, and temporal modeling in a unified architecture, and designing learning paradigms that better handle long-term, irregularly sampled observations. We hope ChronoEarth will serve as a foundation for advancing temporally grounded Earth observation modeling.


\bibliographystyle{plainnat}
\bibliography{reference}


\clearpage
\setcounter{page}{1}
\crefalias{section}{appendix}
\crefname{appendix}{Appendix}{Appendices}
\Crefname{appendix}{Appendix}{Appendices}
\appendix

\section{Additional Related Works}
\textbf{Remote Sensing Representations.}~
Most multispectral geospatial foundation models build upon Vision Transformers (ViT) with modified patch embedding layers to accommodate multi-band inputs~\citep{seco,spectralgpt,croma}. While effective for MSI with around 10 bands, directly projecting high-dimensional hyperspectral inputs into spatial tokens may weaken explicit spectral interaction modeling. 

To better preserve spectral information, SatMAE~\citep{satmae} introduces heuristic channel grouping during patch embedding, and DOFA~\citep{dofa} generates wavelength-conditioned channel weights to adapt spectral parameterization. However, both methods primarily address spectral structure at initialization. Architectures designed specifically for HSI further strengthen spectral interaction modeling. SpectralViT~\citep{spectralearth} performs spectral projection prior to spatial transformer processing, enabling more explicit cross-band relationship modeling. HyperSigma~\citep{hypersigma} separates spectral and spatial attention mechanisms and fuses them before downstream tasks. LESSViT~\citep{less} explicitly models the spatial–spectral interactions through their proposed efficient attention module.

Beyond HSI-specific architectures, large-scale MSI foundation models such as Prithvi~\citep{prithvi}, embedding-oriented geospatial models such as AlphaEarth~\citep{alphaearth} and general-purpose vision foundation models like DINOv3~\citep{dinov3} have demonstrated strong representation learning performance. However, these models are primarily developed for RGB or limited-band MSI and require adaptation to support high-dimensional HSI data.

\textbf{EO-1 Hyperion Mission} The EO-1 Hyperion mission~\citep{EO1H}, launched by NASA in 2000 as part of the Earth Observing-1 program, represents the world’s \textit{longest-operating} spaceborne hyperspectral imaging spectrometer so far (as of 2025), with a mission span exceeding 16 years (2001–2017). Hyperion captured contiguous spectral measurements across more than 240 narrow bands (400–2500 nm) at 30 m spatial resolution, enabling fine-grained analysis of vegetation, soil, and mineral characteristics. Despite its narrow 7.7 km swath and experimental design, EO-1 Hyperion pioneered orbital hyperspectral imaging and provided a cornerstone dataset for subsequent missions such as PRISMA~\citep{PRISMA} (2019-) and EnMAP~\citep{EnMAP} (2021-), which continue to extend its legacy but have not yet achieved comparable temporal longevity.

\section{Dataset Release and Accessibility}~\label{appendix:release}

\subsection{Data Organization.}
ChronoEarth-492K is distributed in two components: hyperspectral imagery (HSI) and benchmark annotations. All samples are indexed by a deterministic spatial identifier (UID) of the form \texttt{[UTM\_zone:column:row]}, which uniquely identifies a fixed cell in the global patch grid.

\subsubsection{Hyperspectral data}
Each hyperspectral sample is stored as a $128\times128\times155$ radiometrically harmonized hyperspectral cube in GeoTIFF format. The GeoTIFF file also contains the associated geospatial metadata. The dataset is organized hierarchically by geographic region and spatial UID. Multiple temporal observations of the same location are stored under the same UID directory and indexed by acquisition timestamp. The timestamp follows the \texttt{YYYYDDD} convention, where \texttt{DDD} denotes the day-of-year:

\begin{center}
\texttt{dataset/<Region>/<UID>/<UID>\_<YYYYDDD>.TIF}
\end{center}

A global metadata table stored at 
\begin{center}
\texttt{dataset/metadata.parquet}
\end{center}

records the file paths and metadata for all hyperspectral samples, enabling efficient indexing and filtering without traversing the
directory structure.

\subsubsection{Benchmark labels}
For each dataset under our benchmark, task-specific metadata files are provided at

\begin{center}
\texttt{benchmark\_labels/<dataset>/<dataset>\_<task\_type>.parquet}
\end{center}

where \texttt{task\_type} denotes the evaluation protocol and can be \texttt{static}, \texttt{sh} (short-horizon), or \texttt{lh} (long-horizon). These metadata tables record the hyperspectral sample path, label path, location UID, acquisition timestamp, and additional task-specific fields. For multi-label classification tasks, the class labels are stored directly in the metadata table as multi-hot vectors. 

The minimum observation threshold can be conveniently applied using the metadata tables. For short-horizon (SH) tasks, the filtering is applied to samples that share the same label, ensuring that each annual label has a sufficient number of temporal observations. For long-horizon (LH) tasks, the filtering is applied to samples associated with the same location UID, ensuring that each location contains enough temporal observations to form a valid long-term sequence.

For dense prediction tasks such as segmentation, label maps are stored as GeoTIFF files at
\begin{center}
\texttt{benchmark\_labels/<dataset>/labels/<UID>/<dataset>\_<year>.tif}
\end{center}
with geospatial metadata preserved in the file headers.

\begin{table}[t]
\centering
\caption{Licensing and providers of external datasets used in ChronoEarth-492K and ChronoEarth-Benchmark.}
\label{tab:dataset_licenses}
\setlength{\tabcolsep}{6pt}
\renewcommand{\arraystretch}{1.15}
\resizebox{\columnwidth}{!}{
\begin{tabular}{ccc}
\toprule
\textbf{Dataset} & \textbf{Provider} & \textbf{License / Usage Terms} \\
\midrule
EO-1 Hyperion~\cite{EO1H} & NASA / USGS EROS\footnotemark[1] & Public Domain \\
CDL~\cite{cdl} & USDA NASS\footnotemark[2] & U.S. Government Data (Public Use) \\
CLCD~\cite{clcd} & Chinese Academy of Sciences\footnotemark[3] & Academic Research Use \\
NLCD~\cite{nlcd} & USGS / MRLC Consortium\footnotemark[4] & Public Domain \\
ISDASoil~\cite{isda} & ISDA / World Agroforestry (ICRAF)\footnotemark[5] & CC-BY 4.0 \\
CORINE~\cite{corine} & European Environment Agency (EEA)\footnotemark[6] & Copernicus Open License \\
GFC~\cite{gfc} & University of Maryland / Google Earth Engine\footnotemark[7] & CC-BY 4.0 \\
\bottomrule
\end{tabular}
}

\end{table}

\footnotetext[1]{EO-1 Hyperion data are archived and distributed by the USGS EROS center and released in the public domain: \url{https://www.usgs.gov/centers/eros/science/usgs-eros-archive-earth-observing-one-eo-1-hyperion}.} 

\footnotetext[2]{USDA NASS Cropland Data Layer: \url{https://www.nass.usda.gov/Research_and_Science/Cropland/}.} 

\footnotetext[3]{China Land Cover Dataset (CLCD): \url{https://doi.org/10.5281/zenodo.5816591}.}

\footnotetext[4]{National Land Cover Database (NLCD): \url{https://www.mrlc.gov/}.}

\footnotetext[5]{ISDASoil Africa: \url{https://www.isda-africa.com/isdasoil/}.}

\footnotetext[6]{CORINE Land Cover: \url{https://land.copernicus.eu/pan-european/corine-land-cover}.}

\footnotetext[7]{Global Forest Change dataset~\cite{gfc}: \url{https://earthenginepartners.appspot.com/science-2013-global-forest}.}

\subsection{Licensing and Ethics.}
ChronoEarth-492K and its benchmarks are constructed from NASA’s EO-1 Hyperion archive~\citep{EO1H} and several publicly available geospatial datasets, including CDL~\citep{cdl}, CLCD~\citep{clcd}, NLCD~\citep{nlcd}, ISDASoil~\citep{isda}, CORINE Land Cover~\citep{corine}, and Global Forest Change (GFC)~\citep{gfc}. All external datasets are used in accordance with their respective licensing terms and
data usage policies. The providers and licensing conditions of these datasets are summarized in Table~\ref{tab:dataset_licenses}. ChronoEarth contains no personally identifiable information, as all data originates from satellite-based Earth observation products. ChronoEarth-492K and the ChronoEarth-Benchmark will be released under an open academic license permitting non-commercial research use.

\subsection{Long-Term Reproducibility.}
To facilitate verification and future extensions, we release the complete data processing pipeline, including scripts for patch extraction, UID-based spatial indexing, temporal sequence construction, and split generation. These tools allow researchers to reproduce the dataset construction process or extend the dataset with additional observations.

\section{ChronoEarth Dataset Construction Details}
\label{appendix:chronoearth}
\subsection{Data Acquisition}
The ChronoEarth dataset was constructed directly from the U.S. Geological Survey (USGS) Earth Resources Observation and Science (EROS) Center archive of Level-1T (terrain-corrected) Hyperion products~\citep{usgs_eo1}. All scenes available between May 1, 2001 and March 13, 2017 were systematically queried and downloaded via the USGS M2M API\footnote{https://m2m.cr.usgs.gov/api/docs/json/}.

\subsection{Spectral Band Filtering and Harmonization}
Since Level-1T products are already orthorectified and radiometrically calibrated by USGS, no additional atmospheric or geometric corrections were applied. Each tile initially contained 242 spectral bands. Following~\citep{process_eo1}, we removed unstable or low-signal bands primarily affected by instrument noise and water-vapor absorption (e.g., around 1400 nm and 1900 nm), as summarized in Table IV of that study. After filtering, 155 spectral bands were retained and spectrally aligned to a consistent wavelength grid to ensure inter-scene uniformity. Each tile was subsequently georeferenced using its metadata, enabling further processing and curation of downstream tasks.

\subsection{Global Spatial Gridding and Patch Extraction}
To construct a spatially and temporally consistent dataset, each Level-1T scene was patched according to its native UTM projection (EPSG:32XXX series, WGS84 datum) as provided in the USGS metadata. For each UTM zone, we established a global gridding system anchored at the $(0, 0)$ origin in projected coordinates. All scenes were resampled to a uniform 30 m pixel size to match the nominal ground sampling distance of Hyperion. Each scene was divided into non-overlapping $128 \times 128$ pixel patches, aligned to the fixed global grid of its corresponding UTM zone. This process assigns every patch a unique and deterministic location identifier (UID) in the format \texttt{[UTM\_zone:column:row]}, enabling unambiguous spatial indexing and retrieval. Patches sharing the same UID but acquired at different timestamps naturally form temporal stacks representing repeated observations of the same location.

Because Hyperion’s narrow swath results in partial coverage and occasional edge gaps, we allowed up to 10\% of nodata pixels per patch to balance spatial diversity with data quality. This patching strategy ensures consistent geodetic alignment within each UTM zone and provides a coherent foundation for long-term temporal analysis across the ChronoEarth corpus.

\begin{table}[t]
\centering
\caption{Distribution of spatial locations and hyperspectral patches across geographic regions in ChronoEarth-492K.}
\setlength{\tabcolsep}{6pt}
\renewcommand{\arraystretch}{1.15}
\resizebox{0.9\columnwidth}{!}{
\begin{tabular}{cccccccccc}
\toprule
 & \textbf{AF} & \textbf{NA} & \textbf{SWA} & \textbf{LA} & \textbf{EA} & \textbf{OC} & \textbf{EU} & \textbf{AC} & \textbf{SEA} \\
\midrule
\#Locations & 26,835 & 40,583 & 34,750 & 22,874 & 22,958 & 13,693 & 12,777 & 5,692 & 5,236 \\
\#Patches & 105,597 & 96,259 & 96,034 & 58,565 & 45,081 & 42,319 & 25,430 & 14,363 & 8,706 \\
\midrule
Total & \multicolumn{9}{r}{\textbf{Locations}: 185,398 \quad \textbf{Patches}: 492,354} \\
\bottomrule
\end{tabular}
}
\label{tab:region_distribution}
\end{table}

\subsection{Regional Distribution} 
To mitigate excessive oceanic coverage and improve geographic diversity, we manually defined nine continental-scale regions when constructing ChronoEarth-492K: Africa (AF), North America (NA), Southwest Asia (SWA), Latin America (LA), East Asia (EA), Oceania (OC), Europe (EU), Arctic (AC), and Southeast Asia (SEA). Table~\ref{tab:region_distribution} summarizes the distribution of spatial locations and hyperspectral patches across these regions. In total, the dataset contains 185,398 unique locations and 492,354 hyperspectral patches. North America, Southwest Asia, and Africa contribute the largest numbers of samples, while smaller but geographically distinct regions such as Oceania, Europe, Southeast Asia, and the Arctic provide additional environmental diversity. This regional partitioning ensures that the dataset captures diverse environmental conditions across continents, climates, and land-cover regimes.

\begin{figure}[t]
    \centering
    \includegraphics[width=\linewidth]{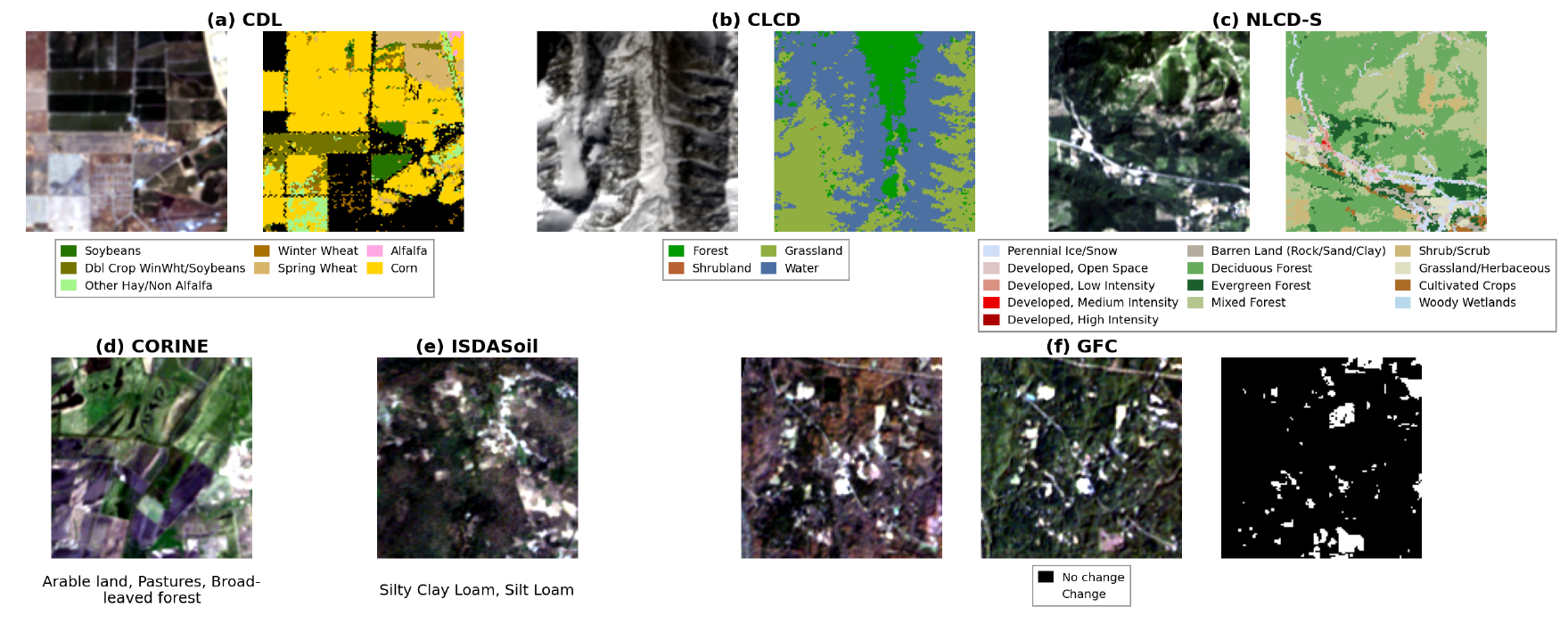}
    \caption{Sample pseudo-RGB images and labels from the ChronoEarth-Benchmark.}
    \label{fig:examples}
\end{figure}

\begin{table*}[t]
\centering
\caption{Benchmark split statistics across datasets and temporal configurations.}
\setlength{\tabcolsep}{6pt}
\renewcommand{\arraystretch}{1.15}

\begin{subtable}{0.32\linewidth}
\centering
\caption{Static Tasks}
\resizebox{\linewidth}{!}{
\begin{tabular}{lccc}
\toprule
\textbf{Dataset} & \textbf{Train} & \textbf{Val} & \textbf{Test} \\
\midrule
CLCD~\citep{clcd} & 12,271 & 1,721 & 3,622 \\
CDL~\citep{cdl} & 2,773 & 526 & 6,863 \\
NLCD-S~\citep{nlcd} & 33,316 & 2,807 & 19,353 \\
ISDASoil~\citep{isda} & 10,620 & 1,204 & 5,984 \\
\bottomrule
\end{tabular}}
\end{subtable}
\hfill
\begin{subtable}{0.32\linewidth}
\centering
\caption{Short-Horizon (SH) Tasks}
\resizebox{\linewidth}{!}{
\begin{tabular}{lccc}
\toprule
\textbf{Dataset} & \textbf{Train} & \textbf{Val} & \textbf{Test} \\
\midrule
CLCD~\citep{clcd} & 517 & 100 & 452 \\
CDL~\citep{cdl} & 1,220 & 182 & 3,439 \\
NLCD-S~\citep{nlcd} & 2,505 & 206 & 1,366 \\
ISDASoil~\citep{isda} & 851 & 78 & 620 \\
\bottomrule
\end{tabular}}
\end{subtable}
\hfill
\begin{subtable}{0.32\linewidth}
\centering
\caption{Long-Horizon (LH) Tasks}
\resizebox{\linewidth}{!}{
\begin{tabular}{lccc}
\toprule
\textbf{Dataset} & \textbf{Train} & \textbf{Val} & \textbf{Test} \\
\midrule
CLCD~\citep{clcd} & 275 & 23 & 162 \\
CDL~\citep{cdl} & 349 & 27 & 162 \\
NLCD-S~\citep{nlcd} & 1,268 & 91 & 453 \\
\bottomrule
\end{tabular}}
\end{subtable}

\vspace{0.6em}

\begin{subtable}{0.6\linewidth}
\centering
\caption{Static CORINE with Distribution Shifts}
\resizebox{\linewidth}{!}{
\begin{tabular}{lcccccc}
\toprule
\textbf{Dataset} & \textbf{Train} & \textbf{Val} & \textbf{ID} & \textbf{S-OOD} & \textbf{T-OOD} & \textbf{ST-OOD} \\
\midrule
CORINE~\citep{corine} & 4,705 & 785 & 2,353 & 6,731 & 2,458 & 2,742 \\
\bottomrule
\end{tabular}}
\end{subtable}

\vspace{0.6em}

\begin{subtable}{0.45\linewidth}
\centering
\caption{GFC with Distribution Shifts}
\resizebox{\linewidth}{!}{
\begin{tabular}{lcccc}
\toprule
\textbf{Dataset} & \textbf{Train} & \textbf{Val} & \textbf{ID-Test} & \textbf{OOD-Test} \\
\midrule
GFC~\citep{gfc} & 1,147 & 231 & 976 & 3,155 \\
\bottomrule
\end{tabular}}
\end{subtable}

\label{tab:benchmark_splits}
\end{table*}

\section{ChronoEarth-Benchmark Datasets Introduction} \label{appendix:downstream_dataset}
In this section, we provide a detailed description of each dataset included in the benchmark. Representative examples from each dataset are shown in \Cref{fig:examples}. Benchmark split statistics are provided in \Cref{tab:benchmark_splits}.

\subsection{CDL: Crop Type Segmentation}
The Cropland Data Layer (CDL)~\citep{cdl} is an annual 30m crop classification product published by the USDA National Agricultural Statistics Service since 2008. It defines up to 254 land-cover and crop categories, including cultivated crops and non-agricultural surfaces. According to USDA validation reports, CDL achieves 85–95\% accuracy for major crops, with reduced reliability for minor or region-specific classes. For this work, we use CDL maps from 2008 to 2017 and align them with EO-1 Hyperion acquisitions according to location and year. To mitigate the severe class imbalance in the original CDL taxonomy while retaining the overall distributional characteristics of major crop types, we retain only crop categories occupying more than 0.1\% of the total area and aggregate the remaining long-tail classes into a single \texttt{[Other Crops]} category. All non-agricultural classes were projected into a single background category. The resulting dataset has 15 agricultural classes and one background class. After patching the projected labels, we further removed patches containing more than 80\% background pixels to ensure effective learning and stable class representation. Following the entropy-based diversity filtering, the final CDL dataset contains 10,162 image–label pairs drawn from 2,864 unique locations and annotated with 4,841 distinct label patches, providing a stable and semantically coherent benchmark for agricultural segmentation. 

\begin{figure}[t]
    \centering
    \includegraphics[width=\linewidth]{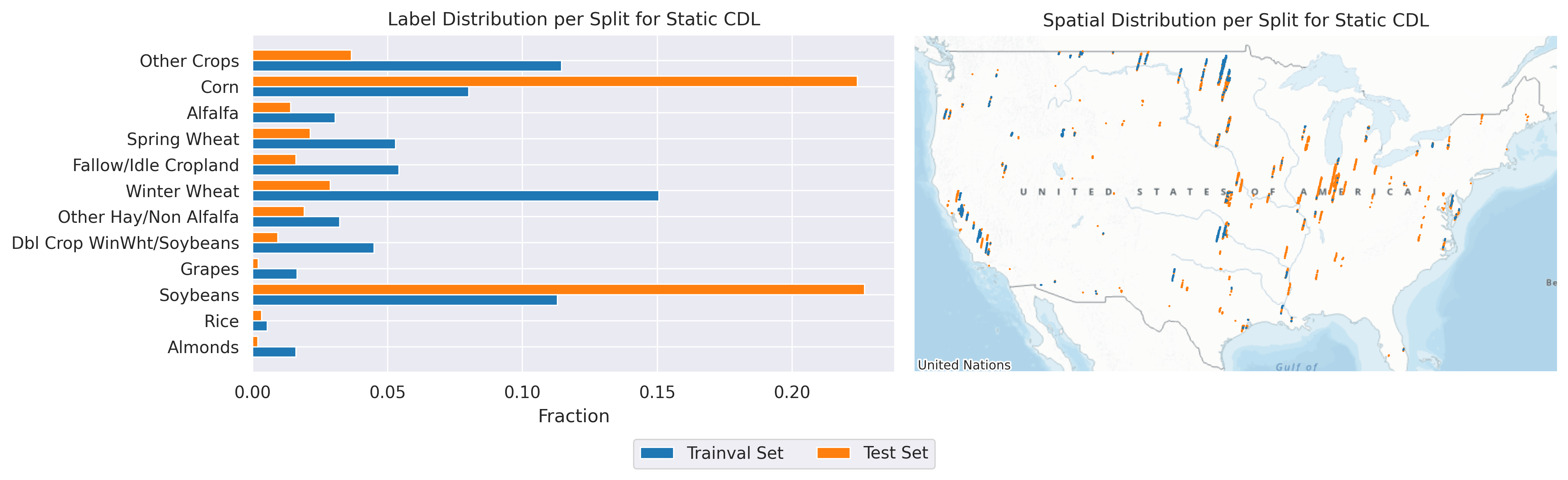}
    \caption{Label and spatial distribution per split for static CDL dataset.}
    \label{fig:cdl_stat}
\end{figure}

We visualize the label distributions of the trainval and test splits for the static CDL, along with their spatial distributions, in \Cref{fig:cdl_stat}. The figure reveals a substantial geographic separation between the trainval and test sets, which results in noticeable differences in their label distributions. This spatial distribution shift makes CDL a particularly challenging task in our benchmark.

\subsection{CLCD: Land Cover Segmentation}
The China Land-Cover Dataset (CLCD)~\citep{clcd} is an annual 30 m land-cover product released through the National Earth System Science Data Center (NESSDC) of China. It provides nationwide land-cover maps from 2001 to 2020, generated from Landsat surface reflectance composites using a random forest classifier trained on national reference samples. CLCD defines nine major land-cover types. Reported overall accuracies range from 76.5\% to 82.5\%, with a mean of approximately 79.3\% across years. For this work, we use CLCD maps from 2001 to 2017. After patching, removing patches with more than 10\% background coverage, and applying entropy-based filtering, the resulting CLCD subset comprises 17,614 image–label pairs from 9,752 unique locations and 12,079 label patches. It serves as a key component of our benchmark, extending land-cover segmentation to the Asian continent and enhancing the spatial diversity of cross-regional evaluation.

\begin{figure}[t]
    \centering
    \includegraphics[width=\linewidth]{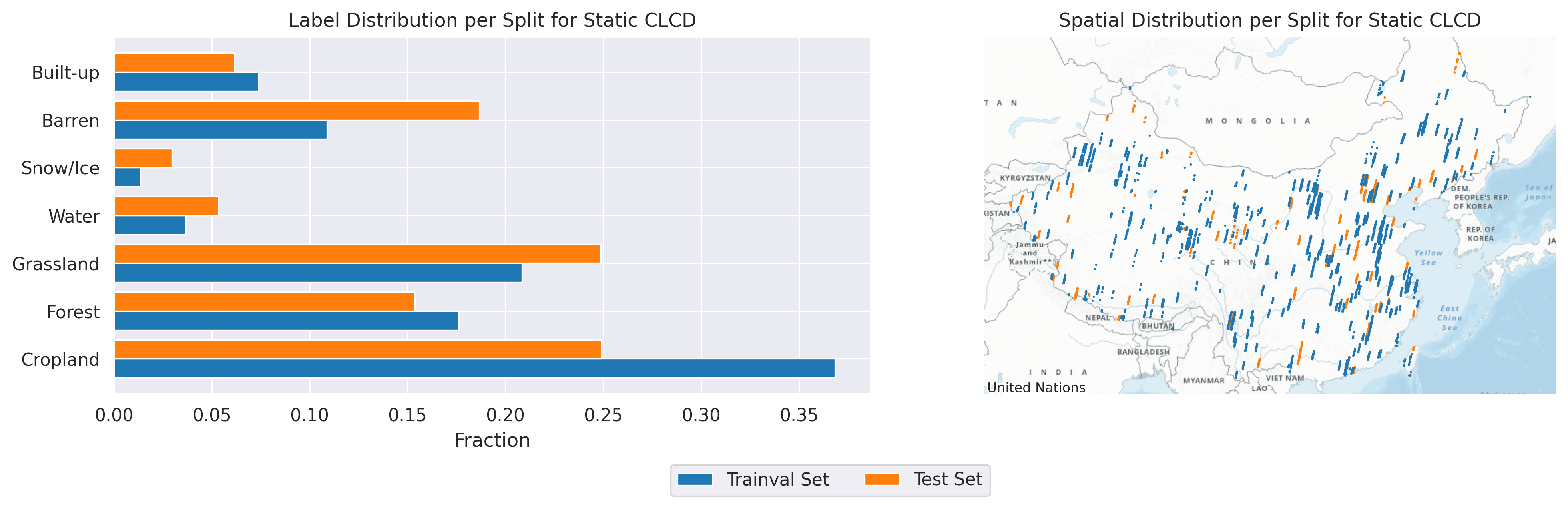}
    \caption{Label and spatial distribution per split for static CLCD dataset.}
    \label{fig:clcd_stat}
\end{figure}

We visualize the label distributions of the trainval and test splits for the static CLCD task, together with their spatial distributions, in \Cref{fig:clcd_stat}. Although the label distribution shift is less pronounced than in CDL, noticeable differences remain. Spatially, the test set contains more samples from western China, whereas the trainval set is more concentrated in eastern China.

\subsection{NLCD: Land Cover Segmentation}
The Annual National Land Cover Database (NLCD)~\citep{nlcd}, developed by the U.S. Geological Survey (USGS) in collaboration with the Multi-Resolution Land Characteristics (MRLC) Consortium, provides a consistent, nationwide 30m land-cover record describing surface composition and dynamics across the conterminous United States. The Collection 1 product suite spans 1985–2024 and includes multiple annual science products capturing land cover, land-use change, and impervious surface properties. From these, we select the land cover component covering 2001–2017 to construct corresponding benchmark tasks.

\subsubsection{Land Cover (NLCD-S).}~The annual NLCD Land Cover product adopts a modified Anderson Level II classification system~\citep{anderson} with 16 thematic classes representing major natural and anthropogenic land-cover types. Reported overall accuracies at Level II range from 78\% to 82\% across epochs, reflecting stable and well-validated mapping quality. We spatially and temporally align the NLCD labels with EO-1 Hyperion observations according to acquisition location and year, remove patches containing more than 10\% background pixels, and apply an entropy-based diversity filter to ensure intra-patch label variety. The resulting dataset comprises 55,476 spatially and temporally aligned image–label pairs from 24,662 unique locations and 35,031 label patches, representing the U.S. component of our land-cover segmentation benchmark.

\begin{figure}[t]
    \centering
    \includegraphics[width=\linewidth]{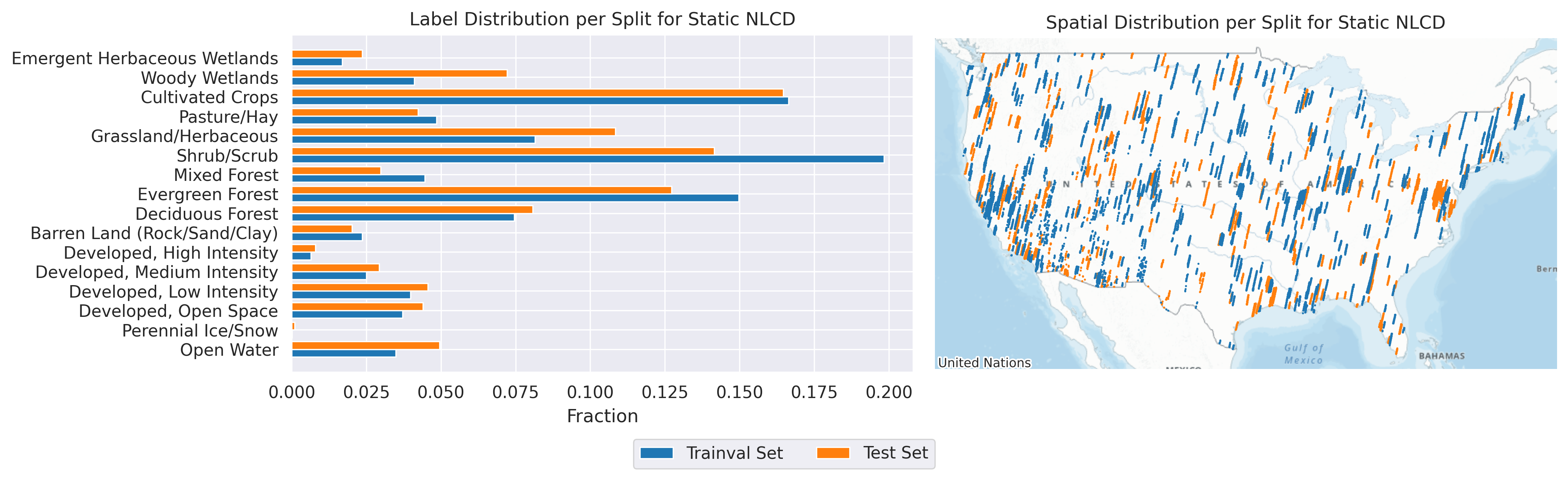}
    \caption{Label and spatial distribution per split for static NLCD-S dataset.}
    \label{fig:nlcd_stat}
\end{figure}

We visualize the label distributions of the trainval and test splits for the static NLCD-S task, together with their spatial distributions, in \Cref{fig:nlcd_stat}. Compared with CDL and CLCD, the label distributions between the trainval and test splits are relatively well matched. The spatial distributions are also more balanced, with both splits covering similar geographic regions.

\subsection{ISDASoil: Soil Texture Fraction Prediction.}
The ISDASoil Africa v1 soil texture map~\citep{isda} provides topsoil (0–20cm) texture information across the African continent at an effective spatial resolution of approximately 250m. The map is generated by machine-learning models that predict continuous sand, silt, and clay fractions and subsequently assign each pixel to one of twelve USDA texture classes. While the continuous fraction predictions has concordance correlation coefficient (CCC) range from 0.78 to 0.85, converting them into discrete texture classes substantially reduces accuracy ($\sim$28–32\%)~\citep{isda_bad}. This reduction reflects the artificial and discontinuous boundaries of the USDA texture diagram, making it an unsuitable ground truth for this benchmark.

To obtain more physically meaningful supervision, we reformulate the task as multi-label soil-texture classification task rather than a categorical segmentation. For each EO-1 Hyperion patch, we annotate it with the set of ISDASoil classes intersecting its footprint. This approach captures spatial heterogeneity within each patch and mitigates the quantization noise introduced by discrete categorical labels. 

Although the ISDASoil product is static rather than annual, we assume that surface soil texture remains stable over the 2001–2017 period of EO-1 Hyperion observations. Accordingly, each Hyperion patch is spatially aligned with the corresponding texture-fraction label based solely on geographic overlap. The resulting dataset comprises 17,808 spatially and temporally aligned image–label pairs from 5,502 unique locations, forming the African soil-texture classification component of our benchmark.

\subsection{CORINE: Crop Type Classification.}
The CORINE Land Cover (CLC) database~\citep{corine}, developed under the Coordination of Information on the Environment (CORINE) program, provides a pan-European inventory of land cover and land use with 44 thematic classes at 100m spatial resolution. Reported overall thematic accuracies exceed 85\%, ensuring high reliability for continental-scale mapping. The CLC product is updated every six years with new status and change layers; for this benchmark, we use the land-cover layers from 2006, 2012, and 2018.

Because the CORINE map has a coarser spatial resolution than the 30m EO-1 Hyperion imagery, we construct a multi-label land-cover classification benchmark in which each Hyperion patch is annotated with the set of CLC classes intersecting its footprint. To ensure temporal consistency, Hyperion acquisitions are aligned to the nearest available CLC epoch: 2001–2006 scenes use CLC 2006, 2007–2012 scenes use CLC 2012, and 2013–2017 scenes use CLC 2018. The resulting dataset comprises 19,774 spatially aligned and temporally associated image–label pairs from 9,706 unique locations and 10,905 unique labels. Following prior work~\citep{spectralearth}, we map the original 44 CLC classes to 19 aggregated categories according to the BigEarthNet taxonomy~\citep{ben}. This dataset extends our benchmark to European landscapes, complementing the NLCD and CLCD counterparts for cross-continental land-cover analysis. 

\subsection{GFC: Forest Loss Detection}
The Global Forest Change (GFC) v1.12 dataset~\citep{gfc} provides annual forest loss and gain at 30m spatial resolution, enabling fine-scale monitoring of global deforestation and reforestation dynamics from 2000 to 2023. The product includes an annual forest loss layer, where each pixel indicates the year that deforestation happened. We utilize this annual forest loss layer to construct a change detection benchmark. For each location with multiple EO-1 Hyperion observations across different years, we examine the corresponding GFC loss-year labels to determine whether a forest-cover change occurred between the two acquisition times. Pairs of observations showing such changes are collected to form the change-detection dataset, while static locations (no detected loss) are excluded. To ensure meaningful signal variation, only patches in which more than 10\% of pixels exhibit change are retained. The resulting dataset com 17,214 samples from 319 unique locations. The resulting dataset comprises 17,214 temporally paired image–label samples from 319 unique locations worldwide. Given the global coverage of GFC, the resulting dataset is organized into regional sub-tasks, each corresponding to a major continental zone (e.g., North America, Latino America, Africa, Europe, East Asia, Oceania and Southwest Asia). This structure allows for both regional fine-tuning and global generalization evaluation of temporal change detection from hyperspectral observations.

\section{ChronoEarth-Benchmark Processing}
\subsection{Entropy-based Diversity Filtering} \label{appendix:filter}
To ensure that each patch contains sufficient semantic diversity, we compute the normalized Shannon entropy of its pixel-wise label distribution. For a given label patch $\mathbf{Y} \in \mathbb{N}^{H \times W}$ with at most $C$ total classes, let $p_c$ denote the empirical frequency of class $c$ within the patch. The normalized entropy is defined as:
\begin{equation}
    H_\mathrm{norm}(\mathbf{Y}) = - \frac{1}{\log C} \sum_{c=1}^{C} p_c \, \log p_c,
\end{equation}
where $p_c = \frac{1}{HW} \sum_{i,j} \mathbb{I}(\mathbf{Y}_{i,j} = c)$ and $\sum_c p_c = 1$. 
A perfectly homogeneous patch yields $H_\mathrm{norm}=0$, while a maximally diverse patch (uniform class distribution) yields $H_\mathrm{norm}=1$.

We retain only patches whose entropy exceeds a predefined threshold $\tau$. By default, we set $\tau=0.1$, which effectively discards semantically trivial or low-diversity regions while preserving information-rich samples for training.

\section{Temporal Module}\label{appendix:temporal}
Given a sequence of $T$ hyperspectral observations, each frame is first encoded independently by the backbone into token features $Z \in \mathbb{R}^{B \times T \times N \times D}$, where $B$ is the batch size, $N$ is the number of spatial tokens, and $D$ is the embedding dimension. We then reshape the tensor to $\mathbb{R}^{(B \cdot N) \times T \times D}$ and apply a temporal attention module along the $T$ dimension, enabling each spatial location to aggregate information across time. Temporal positional embeddings are added to preserve ordering, and a causal mask is applied to prevent information leakage from future frames. During training, the final frame is masked, and the model predicts its representation using the preceding $T-1$ observations. The temporal module outputs features of shape $\mathbb{R}^{B \times N \times D}$, which are passed to a reconstruction decoder to recover pixel-space targets.

To ensure sufficient temporal context, we restrict training to locations with at least three observations, allowing the model to condition on multiple preceding frames. This design isolates temporal reasoning from spatial representation learning, as the backbone is frozen and only the temporal module and decoder are optimized. Notably, this objective is less aligned with classification tasks, where the global \texttt{[CLS]} token is not directly supervised through pixel reconstruction. As a result, tasks such as ISDASoil exhibit limited gains from temporal pretraining in the short-horizon setting.

\section{Hyperparameter Choices for Evaluation}~\label{appendix:hparam}
For all downstream evaluations, we finetune each pretrained model using the AdamW optimizer~\citep{adamw} and sweep the learning rate over a fixed grid $1\mathrm{e}{-5}$, $3\mathrm{e}{-5}$, $5\mathrm{e}{-5}$, $8\mathrm{e}{-5}$, $1\mathrm{e}{-4}$. In addition to the learning rate, we use a batch size of 32 and keep it fixed across all datasets. To ensure comparable optimization across datasets with different sample sizes, we choose the number of finetuning epochs based on the number of training samples so that each experiment receives a sufficient number of optimization steps. During training, we apply a linear warmup over the first $5\%$ of the total steps, followed by cosine annealing for the learning rate schedule, and use a weight decay of $5\times10^{-2}$. \Cref{tab:epochs} summarizes the number of finetuning epochs used for evaluation under the Static, Short-Horizon, and Long-Horizon settings for each dataset.


\begin{table}[t]
\caption{Epochs of finetuning for each dataset on different temporal configurations.}
\centering
\setlength{\tabcolsep}{5pt}
\renewcommand{\arraystretch}{1.15}
\resizebox{0.4\linewidth}{!}{
\begin{tabular}{lccc}
\toprule
\textbf{Dataset} & \textbf{Static} & \textbf{SH} & \textbf{LH} \\
\midrule
CLCD~\citep{clcd}     & 10 & 50 & 50 \\
CDL~\citep{cdl}       & 50 & 50 & 50 \\
NLCD-S~\citep{nlcd}   & 10 & 10 & 50 \\
GFC~\citep{gfc}       & 10 & -- & -- \\
ISDASoil~\citep{isda} & 10 & 50 & -- \\
CORINE~\citep{corine} & 10 & -- & -- \\
\bottomrule
\end{tabular}
}
\label{tab:epochs}
\end{table}


\end{document}